\documentclass[runningheads]{llncs}
\usepackage[T1]{fontenc}
\usepackage{graphicx}
\usepackage{amsmath}
\usepackage{tikz}
\usepackage{amsfonts} 
\usepackage{amssymb}  
\usepackage[numbers]{natbib}
\usepackage{makecell}
\usepackage{url} 
\usepackage[colorlinks=true, linkcolor=blue, citecolor=blue, urlcolor=blue]{hyperref} 
\usepackage{booktabs} 
\usepackage{hhline}
\usepackage{orcidlink}
\usepackage{svg}
\usepackage{xcolor}
\usepackage{arydshln} 
\usepackage[table]{xcolor} 
\usepackage{colortbl} 
\usepackage{multirow}

\usepackage{subcaption}
\begin{document}
\title{An Enhanced Dual Transformer Contrastive Network for Multimodal Sentiment Analysis}
\titlerunning{DTCN for Multimodal Sentiment Analysis}
%
%
\author{Phuong Q. Dao\inst{1}\orcidlink{0009-0005-5010-5307} \and
Mark Roantree\inst{2}\orcidlink{0000-0002-1329-2570} \and
Vuong M. Ngo\inst{1}\orcidlink{0000-0002-8793-0504}\thanks{Corresponding author: \email{vuong.nm@ou.edu.vn}}}
\authorrunning{Phuong Q. Dao et al.}
%
\institute{Ho Chi Minh City Open University, Ho Chi Minh, Vietnam \and Insight Centre for Data Analytics \& School of Computing, Dublin City University, Dublin, Ireland\\
}
\maketitle 
\begin{abstract}
Multimodal Sentiment Analysis (MSA) seeks to understand human emotions by jointly analyzing data from multiple modalities—typi\-cally text and images—offering a richer and more accurate interpretation than unimodal approaches. In this paper, we first propose BERT-ViT-EF, a novel model that combines powerful Transformer-based encoders—BERT for textual input and ViT for visual input—through an early fusion strategy. This approach facilitates deeper cross-modal interactions and more effective joint representation learning. To further enhance the model’s capability, we propose an extension called the Dual Transformer Contrastive Network (DTCN), which builds upon BERT-ViT-EF. DTCN incorporates an additional Transformer encoder layer after BERT to refine textual context (before fusion) and employs contrastive learning to align text and image representations, fostering robust multimodal feature learning.
Empirical results on two widely used MSA benchmarks—MVSA-Single and TumEmo—demonstrate the effectiveness of our approach. DTCN achieves best accuracy (78.4\%) and F1-score (78.3\%) on TumEmo, and delivers competitive performance on MVSA-Single, with 76.6\% accuracy and 75.9\% F1-score. These improvements highlight the benefits of early fusion and deeper contextual modeling in Transformer-based multimodal sentiment analysis.

\end{abstract}

\section{Introduction}
In the age of social media and ubiquitous user-generated content, understanding human sentiment has become increasingly important. As a result, Multimodal Sentiment Analysis (MSA) has emerged as a critical area in natural language processing and computer vision. Unlike traditional text-only approaches, MSA leverages complementary signals across modalities to better capture subtle emotional expressions, sarcasm, or visual context that may not be apparent in language alone. This capability is particularly vital in domains such as social media monitoring, customer feedback analysis, and affective computing, where emotions are inherently multimodal. Consequently, developing effective fusion strategies and modality-aware representations is essential for advancing sentiment understanding in real-world, noisy, and richly contextual environments. Unlike traditional sentiment analysis, which usually relies on textual information alone, MSA incorporates multiple sources of information, including text and images, to provide a richer and more robust understanding of emotional content.

Early approaches to MSA were often based on shallow feature engineering or classical machine learning techniques, treating each modality in isolation \cite{02_Tembhurne2021}. While these methods offered initial insights, they lacked the capacity to capture the deep interdependencies between modalities. With the advent of deep learning, especially Transformer-based architectures, the field has witnessed a significant leap forward \cite{04_yang2021mvan,10_zhu2023skeafn}.

Models such as BERT \cite{devlin-etal-2019-bert,Ngo.APWG.2024} have revolutionized textual understanding by capturing contextual nuances at the word and sentence levels. Similarly, the introduction of the Vision Transformer (ViT) \cite{dosovitskiy2021image} has made it possible to learn hierarchical visual features without relying on convolutional operations. However, integrating these powerful unimodal encoders into a unified multimodal framework remains a challenge. Most existing works adopt a \textit{late fusion} strategy, where modality-specific predictions are combined at the decision level. Although straightforward, this approach often fails to fully exploit cross-modal interactions \cite{06_li-etal-2022-clmlf}.

To address this limitation, recent studies have explored \textit{early fusion} techniques, where representations from different modalities are integrated prior to classification \cite{14_ali2023xai}. Early fusion has been shown to enhance performance by enabling deeper semantic alignment between text and visual content \cite{07_jia-etal-2022-beyond, 26_Wang2024CoAttention}. Furthermore, dual-branch or two-stream architectures have proven effective in retaining the integrity of modality-specific encoders while allowing shared representation learning \cite{13_Ghosh2023}.

In this paper, we present a Dual Transformer Contrastive Network (DTCN) model that combines BERT for text and ViT for images within an early fusion framework. By integrating the [CLS] tokens from each modality at the representation level, our model encourages interaction between modalities at a deeper level. DTCN is evaluated on two widely used benchmark datasets: MVSA-Single and TumEmo, which are standard for assessing MSA systems \cite{23_Chen2024Holistic, 06_li-etal-2022-clmlf, 09_sun2023debiasing, 08_wu2023hgbert, 04_yang2021mvan, 21_Zhang2024CTMWA, 11_zhu2023mulser, 12_zhu2023itin}.

Our main contributions are summarized as follows:
\vspace{-3mm}
\begin{itemize}
    \item We empirically demonstrate that early fusion outperforms late fusion on two benchmark datasets.
    \item BERT-ViT-EF, integrating BERT and ViT using early fusion, is proposed to enable joint cross-modal learning and demonstrates strong performance on MSA tasks. BERT captures contextual textual features, while ViT provides rich visual representations.
    \item DTCN, an enhanced architecture built upon BERT-ViT-EF, incorporates an additional Transformer encoder layer after BERT (before fusion) and employs contrastive learning. This design fosters deeper cross-modal interaction and enhances joint representation learning.
    \item Our models are evaluated and compared against both classical and Transformer-based MSA models on the TumEmo and MVSA-Single datasets, achieving state-of-the-art performance on TumEmo and competitive results on MVSA-Single.
\end{itemize}

The rest of this paper is organized as follows. 
Section~\ref{sec:related_work} reviews relevant research in MSA, focusing on Transformer-based architectures and fusion strategies. 
Section~\ref{sec:proposed_model} presents our proposed Dual Transformer Contrastive Network, detailing the architecture, early fusion mechanism, and training objectives. 
Section~\ref{sec:experiments} describes the datasets, experimental setup, and implementation details. 
Section~\ref{sec:results} reports and analyzes experimental results, including comparisons with existing baselines. 
Finally, Section~\ref{sec:conclusion} concludes the paper and discusses directions for future work.

\vspace{-3mm}
\section{Related Work}
\label{sec:related_work}
\vspace{-2mm}
\subsection{Deep learning and Fusion Strategies for MSA}
MSA combines textual and visual information to better capture emotional signals in user-generated content. Traditional methods often relied on handcrafted features and classical machine learning algorithms \cite{02_Tembhurne2021}, which lacked the capacity to model complex interactions between modalities.

Deep learning approaches marked a significant turning point in MSA. Models based on Convolutional Neural Network (CNN) and Long Short-Term Memory (LSTM), such as Multi-view Attentional Network (MVAN) \cite{04_yang2021mvan} and Sentiment Knowledge Enhanced Attention Fusion Network (SKEAFN) \cite{10_zhu2023skeafn}, incorporated attention mechanisms to better align textual and visual information. However, these models primarily used static embeddings (e.g., GloVe) and ResNet for image features, limiting their contextual understanding.

Fusion strategy remains a critical design factor in MSA \cite{MSASurvey2024}. While many systems adopt late fusion for simplicity, studies like CLMLF \cite{06_li-etal-2022-clmlf} show that joint representation learning via carefully designed loss functions can yield better results. However, CLMLF still combines modalities at a late stage, limiting early interaction. In contrast, early fusion enables joint learning from the start. Intermediate fusion offers a middle ground by integrating modalities at multiple stages after initial unimodal encoding, aiming to balance specialization and interaction.

\vspace{-2mm}
\subsection{Classical MSA Models}
\vspace{-1mm}
\texttt{HSAN-M} \cite{Xu2017HSAN-M} is a hierarchical semantic attentional network for MSA. It leverages image captions as semantic information  and uses attention mechanisms to extract and combine features from different modalities. It aims to capture complementary information by focusing on salient parts of each modality. 

\texttt{MultiSentiNet-M} \cite{Xu2017mvsa} is a deep semantic network that extracts deep semantic features from images using salient detectors and employs a visual feature-guided attention LSTM model to extract important words for sentiment analysis. This model highlights the importance of both visual and textual content for understanding human sentiments. It also leverages scene and object features.

\texttt{Co-Memory-M} \cite{Xu2018Co-MemoryNetwork} utilizes co-memory networks to iteratively model mutual influences between image and text data for MSA. It explores the interrelations between image and text to achieve a more comprehensive understanding of sentiment.

\texttt{Se-MLNN} \cite{Che2021Se-MLNN} is a model that integrates pre-trained image features with contextual text features for visual-textual sentiment analysis. It uses pre-trained encoders for various visual features including objects, scenes, facial expressions, and CLIP embeddings.

\texttt{MGNNS} \cite{yang-etal-2021-multimodal} utilizes a multi-channel graph neural network to integrate heterogeneous multimodal data and learn complex associations. It incorporates a multi-head self-attention mechanism to capture internal data vector correlations, potentially aligning and fusing text tokens and image patches more effectively.\vspace{3mm}

However, classical models often rely on handcrafted features and modular attention mechanisms. While these models can capture basic associations between visual and textual modalities, they struggle to model deeper contextual dependencies—particularly in informal, ambiguous, or noisy data commonly found in real-world social media. Their limited capacity to understand nuanced sentiment expressions, sarcasm, or multimodal irony restricts their overall performance and generalizability. Consequently, there is a growing need for more robust architectures that can learn richer cross-modal representations in an end-to-end manner.

\vspace{-3mm}
\subsection{Transformer-based MSA models}
\vspace{-2mm}

\texttt{MVAN-M} \cite{36_Tameemi2023mvan} is a multimodal emotion classification model that uses a Multi-view Attentional Network. It includes three stages: feature mapping (object/scene for images, local/long-term for text), interactive learning (with a memory network for intra- and cross-view features), and deep feature fusion. The model also supports interpretability by highlighting influential visual and textual cues.

\texttt{CLMLF} \cite{06_li-etal-2022-clmlf} a novel method for multimodal sentiment detection that specifically focuses on text and image data. It encodes text and images into hidden representations and then uses a multi-layer fusion module, built on a Transformer-Encoder, to align and fuse these token-level features, leveraging multi-headed self-attention. CLMLF also incorporates Label-Based Contrastive Learning (LBCL) and Data-Based Contrastive Learning (DBCL) tasks to help the model learn common sentiment-related features within multimodal data, enhancing robustness and improving performance.

\texttt{CTMWA} \cite{21_Zhang2024CTMWA} addresses MSA by focusing on modality-independent emotional cues and robustness to missing modalities. It uses a crossmodal translation encoder to learn shared representations and reconstruct missing ones. A meta-learning-based unimodal weight adaptation strategy further adjusts the loss contribution of each modality dynamically.

\texttt{VSA-PF} \cite{23_Chen2024Holistic} is a robust MSA method that leverages diverse pre-trained models. It includes four components: a visual-textual branch (Swin Transformer + BERTweet), a visual expert branch (face, object, scene, OCR), a CLIP branch for cross-modal alignment, and a BERT-based fusion network.

Although transformer-based models address MSA using attention mechanisms and contrastive learning, they still rely on late fusion strategies. In this approach, features from each modality are processed separately and only integrated at a later stage, which limits the opportunity for early and meaningful interaction between textual and visual features. Furthermore, the models do not fully exploit the strength of powerful Transformer encoders for both modalities in a unified framework. Notably, textual information—which often carries the most sentiment—may not be sufficiently enriched, as it is typically not refined through deeper Transformer-based processing.

To address these limitations, we propose DTCN, an enhanced Dual Transformer Contrastive Network that integrates BERT and ViT through an early fusion strategy. To further strengthen textual representation, we introduce an additional Transformer encoder layer after BERT, allowing for deeper modeling of contextual information. Contrastive learning is applied to align multimodal features and enhance their cross-modal interaction. These design choices enable the model to learn richer cross-modal representations and improve generalization performance on sentiment analysis datasets.

\section{Dual Transformer Contrastive Network Model}
\label{sec:proposed_model}
\vspace{-2mm}

This section introduces the architecture of the proposed DTCN for MSA. The model leverages BERT \cite{devlin-etal-2019-bert} for textual encoding and ViT \cite{dosovitskiy2021image} for visual encoding. \texttt{Early fusion strategy} is deployed to combine modality representations prior to classification, facilitating more effective cross-modal learning, as early fusion has been shown to outperform late fusion in prior work \cite{04_yang2021mvan}.

\subsection{Transformer-Based Architectures}
\vspace{-2mm}
Transformers have recently gained popularity in MSA due to their scalability and strong representational capacity. BERT remains the dominant text encoder in multimodal pipelines. For vision, ViT offers a powerful alternative to CNNs by modeling global context using self-attention.

\begin{figure}
  \centering
  \includegraphics[width=.79\textwidth, height=8.9cm]{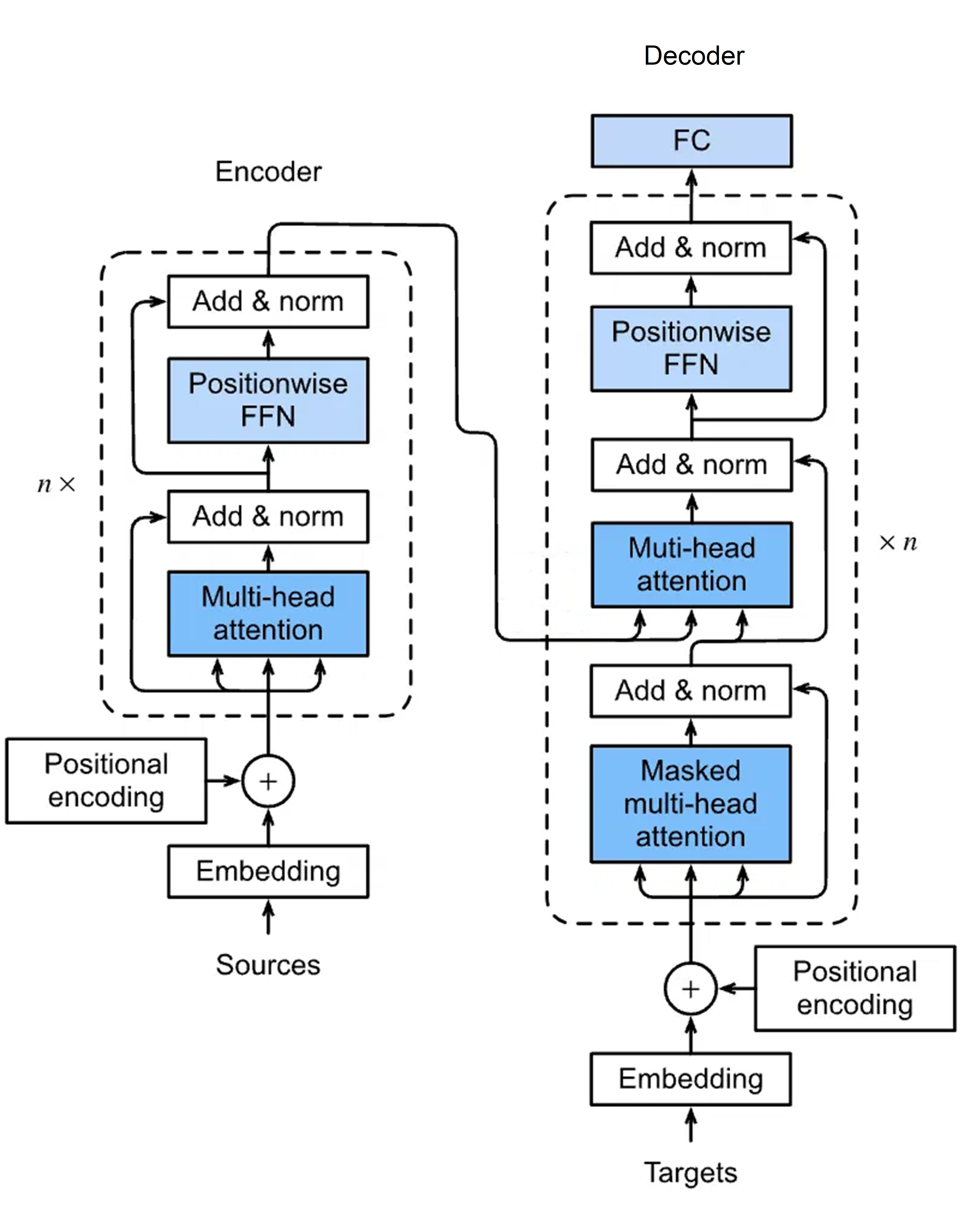} 
  \caption{Transformer Encoder - Decoder Architecture.}
  \label{fig:Transformer_architecture}
\end{figure}

The Transformer architecture showed in Figure~\ref{fig:Transformer_architecture}, as originally introduced by Vaswani et al in \cite{Vaswani2017Transformer}, follows an encoder–decoder structure where both components are composed of repeated layers of multi-head self-attention and position-wise feedforward networks. Each sub-layer is wrapped with residual connections followed by layer normalization, enabling stable training of deep stacks.

In the encoder, each layer allows every token to attend to all others, capturing long-range dependencies efficiently with a constant number of sequential operations per layer. This structure is particularly advantageous for learning complex patterns in sequences, as deeper stacks of encoder layers can hierarchically model higher-order semantic interactions.

In our model, we build upon this architecture by increasing the number of Transformer encoder layers, specifically, within the text branch. This deep stacking enables the model to learn richer and more abstract modality-specific representations before fusion, significantly enhancing performance on challenging MSA benchmarks.

\vspace{-5mm}
\vspace{-5mm}

\begin{figure}
  \centering
  \includegraphics[width=12.5cm]{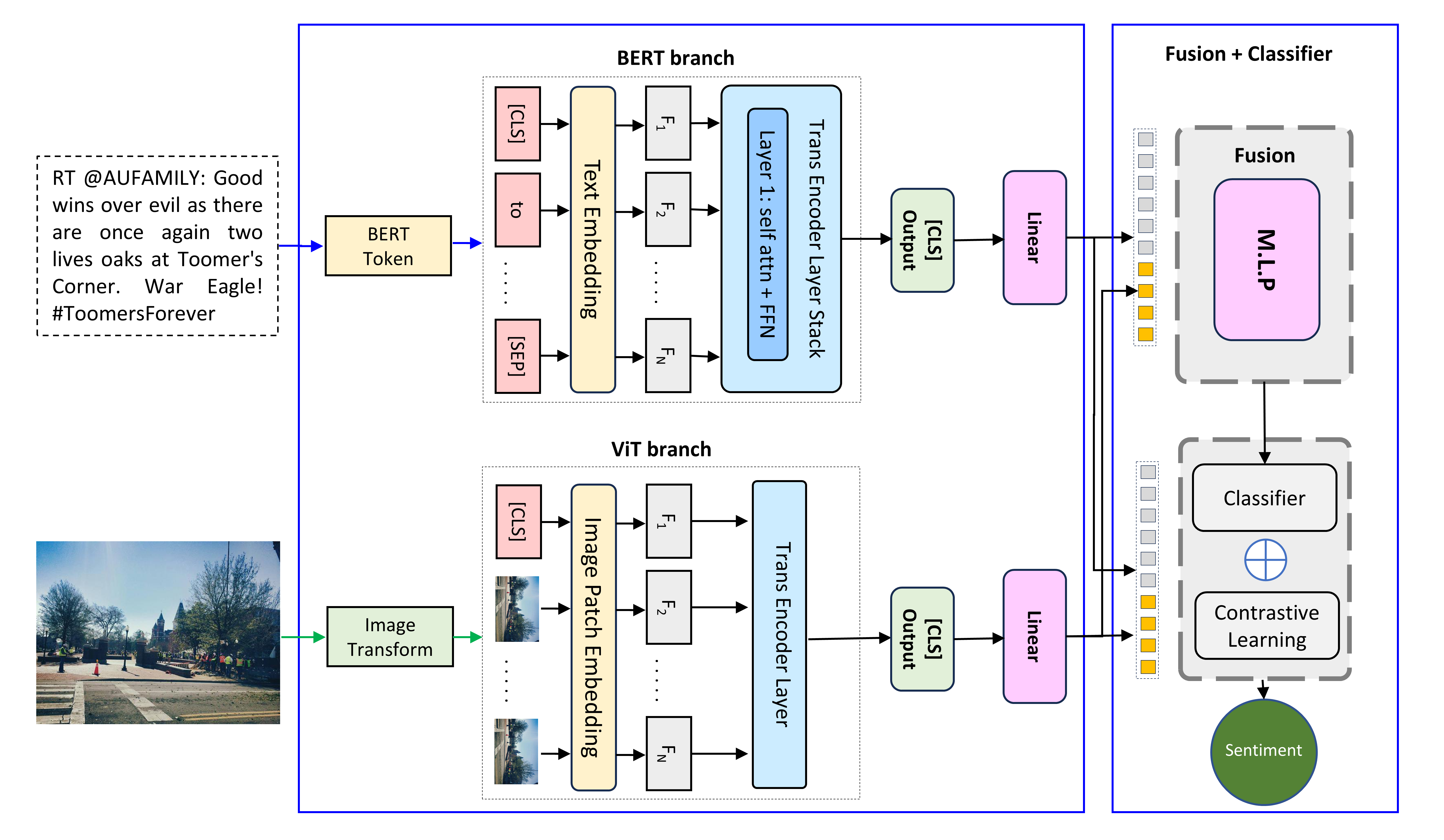} 
  \caption{DTCN Architecture for Image and Text Sentiment Classification.}
  \label{fig:DTCNarchitecture}
\end{figure}

An overview of the DTCN architecture is illustrated in Figure~\ref{fig:DTCNarchitecture} which comprises two modality-specific branches—text and image—followed by a unified fusion and classification module. The BERT branch tokenizes input text and passes it through a pretrained BERT encoder. To enhance semantic abstraction, an additional Transformer encoder layer is stacked on top of the BERT output, further refining the contextual [CLS] representation. In parallel, the ViT branch transforms the input image into patch embeddings and processes them through a pretrained Vision Transformer; its [CLS] token serves as a global visual descriptor.

Both [CLS] vectors are linearly projected and combined using a Multilayer Perceptron (MLP) to form a fused representation. This joint embedding is then passed to a classifier for sentiment prediction. Additionally, a contrastive learning module is employed to align cross-modal semantics by minimizing the distance between matching text–image pairs. The extra transformer layer in the text branch improves modality-specific modeling before fusion, leading to better multimodal sentiment understanding.

\subsection{BERT branch}
The input text is tokenized and passed through a pretrained language model, BERT~\cite{devlin-etal-2019-bert}, which converts each token into a contextualized vector representation. Given a text sequence $T = \{w_1, w_2, ..., w_n\}$, the BERT encoder outputs hidden states: $
H_T^{(0)} = \text{BERT}(T) \in \mathbb{R}^{n \times d}
$. To enhance the contextual representation, we apply a stack of $L_t$ transformer encoder layers over the output of BERT. These additional self-attention layers allow the model to further refine the semantic information relevant to the downstream task: $
H_T = \text{TransformerEncoder}_{\text{text}}(H_T^{(0)}) \in \mathbb{R}^{n \times d}
$. The final sentence-level representation is then extracted using the hidden state corresponding to the special classification token \texttt{[CLS]} at the first position: $
h_T = H_T[0] \in R^{d}$. This global textual representation is passed through a projection layer before being integrated with the visual modality.

\subsection{ViT branch}
Each input image is resized and normalized before being processed by a pretrained ViT model~\cite{dosovitskiy2021image}. The image $I \in R^{H \times W \times 3}$ is first divided into non-overlapping patches, which are then linearly projected and embedded into a sequence of visual tokens:  $
H_I = \text{ViT}(I) \in R^{p \times d}
$. Following standard ViT design, a special classification token \texttt{[CLS]} is prepended to the patch sequence. We use the output representation corresponding to this token as the global image embedding: $
h_I = H_I[0] \in R^{d}$. This global visual representation is projected via a linear layer and then fused with the text representation for downstream tasks.

\subsection{Fusion and Classifier Module}
After extracting modality-specific representations $h_T, h_I \in R^{d}$ from the \texttt{[CLS]} tokens of the text and image encoders, respectively, each is projected through a modality-specific linear layer: $
z_T = \text{Linear}_T(h_T), \quad z_I = \text{Linear}_I(h_I)
$. The final joint representation is obtained by averaging the two projected vectors: $
h_{\text{joint}} = \frac{1}{2} \left( z_T + z_I \right)$. This representation is used for both classification and contrastive learning.

\subsubsection{Contrastive Learning Objective:}
To align representations from different modalities, we use a contrastive learning objective based on the NT-Xent loss~\cite{Chen2020SimCLR}. For a batch of $B$ paired samples, the projected features are first $L_2$-normalized:
\[
\tilde{z}_T = \frac{z_T}{\|z_T\|}, \quad \tilde{z}_I = \frac{z_I}{\|z_I\|}
\] 

We concatenate the features to form a joint representation matrix: \[Z = \begin{bmatrix} \tilde{z}_T \\ \tilde{z}_I \end{bmatrix} \in R^{2B \times d}\]

The similarity between two vectors is computed as: $
S_{ij} = \frac{Z_i^\top Z_j}{\tau}$,
where $\tau$ is a temperature scaling factor. The NT-Xent loss is defined as: $
\mathcal{L}_{\text{contrast}} = \frac{1}{2B} \sum_{i=1}^{B} \left[ \ell(i, i+B) + \ell(i+B, i) \right]
$, 
where $\ell(i, j)$ is the cross-entropy loss with $j$ as the positive index for $i$ and all others as negatives.
\subsubsection{Classification Head:} The fused representation $h_{\text{joint}}$ is passed through a multi-layer perceptron with layer normalization, GELU activation, dropout, and a final linear layer. The prediction is: $
\hat{y} = \text{Softmax}(W \cdot h_{\text{joint}} + b)$. This output is used to compute the standard classification loss $\mathcal{L}_{\text{cls}}$.

\subsubsection{Overall Training Objective:} The model is trained with a joint objective combining classification and contrastive loss: $
\mathcal{L} = \mathcal{L}_{\text{cls}} + \lambda \mathcal{L}_{\text{contrast}}$. Where $\lambda$ is a balancing coefficient controlling the contribution of the contrastive objective.

\section{Experiments}
\label{sec:experiments}

\subsection{Datasets and Preprocessing}
The DTCN model is evaluated using two public multimodal sentiment datasets: MVSA-Single and TumEmo. MVSA-Single contains 5,129 image-text pairs from Twitter, each labeled by one annotator with sentiment (positive, neutral, or negative) for both image and text. Following~\cite{Xu2017mvsa}, we remove pairs with inconsistent image and text labels. If one label is neutral and the other is positive or negative, the non-neutral sentiment is assigned. Pairs with completely contradictory labels (e.g., positive vs. negative) are discarded. After this filtering process, the MVSA-Single dataset contains 4,511 valid image-text pairs. The final dataset statistics are shown in Table~\ref{tab:mvsa_tumemo_stats}(a).

\begin{table}[ht]
\centering
\caption{Statistics of the MVSA and TumEmo Datasets}
\label{tab:mvsa_tumemo_stats}
\vspace{0.6cm}

(a) MVSA dataset (4,511 pairs in 3 classes) \\
\begin{tabular}{p{18mm}p{14mm}p{14mm}p{14mm}}
\toprule
\textbf{Sentiment} & Positive & Neutral & Negative \\
\midrule
\textbf{Number}  & 2,683 & 470  & 1,358 \\
\bottomrule
\end{tabular}

\vspace{1em}

(b) TumEmo dataset (195,265 pairs in 7 classes)\\
\begin{tabular}{p{16mm}p{12mm}p{12mm}p{12mm}p{12mm}p{12mm}p{12mm}p{12mm}}
\toprule
\textbf{Emotion} & Angry & Bored & Calm & Fear & Happy & Love & Sad \\
\midrule
\textbf{Number}  & 14,544 & 32,283 & 18,109 & 20,264 & 50,267 & 34,511 & 25,277 \\
\bottomrule
\end{tabular}

\end{table}

TumEmo \cite{04_yang2021mvan} is comprised of 195,265 image-text pairs, which are each tagged by one of seven emotion labels. This dataset is a large-scale multimodal dataset constructed by crawling image–text posts from Tumblr\footnote{ \url{http://tumblr.com}} to support emotion classification. Unlike previous datasets, TumEmo provides weak emotion annotations derived from user-generated tags, which often reflect the posters’ emotional states. To avoid the high cost of manual labeling, the authors employed a distant supervision approach, treating these user tags as weak labels for emotion classification. The number of image-text pairs for each emotion is shown in Table~\ref{tab:mvsa_tumemo_stats}(b). Both datasets show label imbalance: neutral is underrepresented in MVSA-Single, while TumEmo is skewed toward “Happy”.

\textbf{Preprocessing.} Text data are normalized by replacing emojis, user mentions, and URLs with special tokens. Sentiment-related hashtags (e.g., \texttt{\#happy}, \texttt{\#sad}, \texttt{\#angry}) are removed to prevent label leakage. Additional cleaning includes punctuation simplification and abbreviation standardization, following BERTweet-style normalization.

\textbf{Data Splitting.} All datasets are randomly divided into training, validation, and test sets using an 8:1:1 stratified split.

\subsection{Implementation Details}
BERT-base and ViT-base were employed as backbone encoders, both of which are fine-tuned end-to-end. The experimental setup is as follows:
\begin{itemize}
\item \textbf{Encoder Layers:} A single Transformer encoder layer with 8 attention heads is deliberately added after BERT, increasing the total number of layers in the Bert branch to 13. In contrast, no additional layers are added after ViT, which retains its original 12-layer architecture in the ViT branch.

\item \textbf{Fusion Strategy:} Early fusion via averaging of projected BERT and ViT \texttt{[CLS]} token representations.

\item \textbf{Representation:} \texttt{[CLS]} token is used as the global embedding for both text and image modalities.

\item  \textbf{Classifier:} Multi-layer perceptron with layer normalization, GELU activation, dropout, and a final linear layer.

\item \textbf{Optimization:} Adam optimizer with a learning rate of $2 \times 10^{-5}$, batch size of 16, trained for up to 10 epochs.

\item \textbf{Loss Function:} Cross-Entropy Loss combined with a Contrastive Loss (NT-Xent) with a weight of 0.2.

\item \textbf{Evaluation:} Early stopping based on validation F1-score; best validation and test performance are reported.
\end{itemize}

\section{Results and Analysis}
\label{sec:results}
\subsection{BERT-ViT-EF Evaluation}
We compare early and late fusion strategies employing the same encoder architectures, using Accuracy and F1-score as evaluation metrics \cite{helmer2015similarity, ngo2021structural, ngo2025graph}. As shown in Table~\ref{tab:results_eflf}, \texttt{BERT-ViT-EF (Ours)} achieves the highest performance across both MVSA-S and TumEmo, with an F1-score of 73.3\% and 77.4\%, respectively. Compared to its late fusion counterpart, \texttt{BERT-ViT-LF}, it yields consistent improvements of $+2.9\%$ (MVSA-S) and $+0.4\%$ (TumEmo) in F1-score, confirming the effectiveness of early cross-modal interactions.

\vspace{-5mm}
\begin{table}[ht]
\centering
\caption{Performance Comparison: Early vs Late Fusion}
\begin{tabular}{l|cccc|cccc}
\hline
\multirow{2}{*}{\textbf{Model}} & \multicolumn{4}{c|}{\textbf{MVSA-S}} & \multicolumn{4}{c}{\textbf{TumEmo}} \\
\cline{2-9}
& \textbf{Acc \%} & \scriptsize{(+/-)} & \textbf{F1 \%} & \scriptsize{(+/-)} & \textbf{Acc \%} & \scriptsize{(+/-)} & \textbf{F1 \%} & \scriptsize{(+/-)} \\
\hline
BERT-ResNet50-LF & 66.2 & {\scriptsize (+8.1)} & 64.9 & {\scriptsize (+8.4)} & 76.2 & {\scriptsize (+1.2)} & 76.1 & {\scriptsize (+1.3)} \\
BERT-ResNet50-EF & 72.4 & {\scriptsize (+1.9)} & 71.1 & {\scriptsize (+2.2)} & 76.7 & {\scriptsize (+0.7)} & 76.6 & {\scriptsize (+0.8)} \\
\hdashline
BERT-ViT-LF & 72.6 & {\scriptsize (+1.7)} & 70.4 & {\scriptsize (+2.9)} & 77.1 & {\scriptsize (+0.3)} & 77.0 & {\scriptsize (+0.4)} \\
\textbf{BERT-ViT-EF (Ours)}     & \textbf{74.3} & & \textbf{73.3} & & \textbf{77.4} & & \textbf{77.4} &\\
\hline
\end{tabular}
\label{tab:results_eflf}
\end{table}
\vspace{-3mm}

Furthermore, when comparing across encoder backbones, models based on BERT and ViT outperform those using traditional architectures such as ResNet50. ViT outperforms ResNet in this context as it captures global visual dependencies more effectively through self-attention, which aligns better with textual semantics in cross-modal tasks. These results demonstrate that both early fusion and strong modality-specific encoders are key to improving MSA.

\subsection{DTCN Evaluation}

\begin{figure}
  \centering
  \includegraphics[width=1\textwidth, height=5cm]{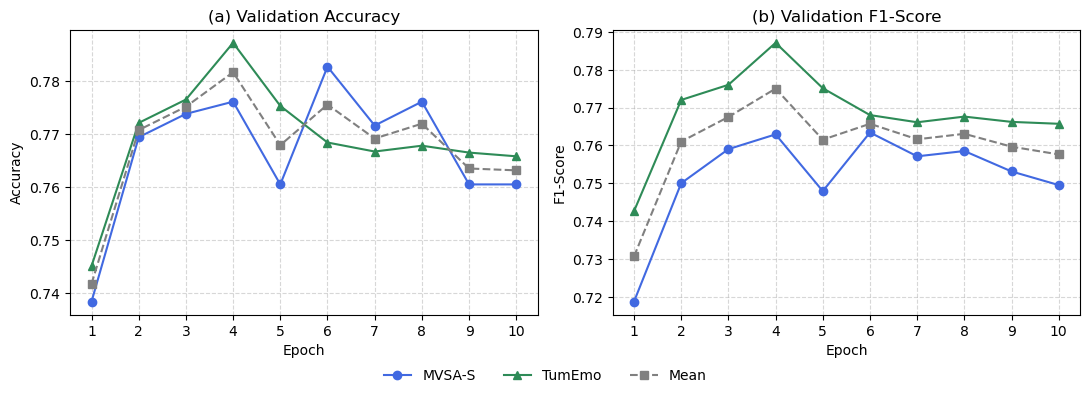} 
  \caption{Validation accuracy and F1-score on MVSA-S and TumEmo over epochs}
  \label{fig:ValAccF1}
\end{figure}

Figure~\ref{fig:ValAccF1} shows the validation accuracy and F1-score over 10 epochs on MVSA-S and TumEmo. The model rapidly improves in the first few epochs, reaching peak performance around epoch 4, as indicated by the mean curve. After that, performance stabilizes with minor fluctuations, suggesting that the model efficiently captures emotional cues early during training and generalizes well across datasets in the MSA setting.

The proposed DTCN model achieves state-of-the-art results on both the MVSA-S and TumEmo datasets, outperforming a range of existing classical and Transformer-based MSA methods discussed in Section \ref{sec:related_work}. As shown in Table~\ref{tab:results_baseline_transform}, DTCN obtains an accuracy of 76.6\% and F1-score of 75.9\% on MVSA-S, while achieving 78.4\% accuracy and 78.3\% F1 on TumEmo, marking the best overall performance across both benchmarks. These improvements can be directly attributed to three key architectural contributions.

First, the early fusion of BERT and ViT allows the model to learn joint representations from the  beginning of the network, fostering deeper cross-modal interactions. Unlike late fusion methods which combine features after separate processing streams, early fusion in DTCN encourages the model to align and interpret sentiment signals from both modalities simultaneously, leading to more coherent and expressive representations.

\vspace{-3mm}
\vspace{-5mm}

\vspace{-6mm}
\begin{table}[ht]
\footnotesize
\centering
\caption{Comparison of DTCN Model against Existing State of the Art}
\begin{tabular}{l|cccc|cccc}
\hline
\multirow{2}{*}{\textbf{Model}} & \multicolumn{4}{c|}{\textbf{MVSA-S}} & \multicolumn{4}{c}{\textbf{TumEmo}} \\
\cline{2-9}
& \textbf{Acc \%} & \scriptsize{(+/-)} & \textbf{F1 \%} & \scriptsize{(+/-)} & \textbf{Acc \%} & \scriptsize{(+/-)} & \textbf{F1 \%} & \scriptsize{(+/-)} \\
\hline
HSAN-M\cite{Xu2017HSAN-M} & 69.9 & {\scriptsize (+6.7)} & 64.6 & {\scriptsize (+11.3)} & 63.1 & {\scriptsize (+15.3)} & 54.0 & {\scriptsize (+24.3)} \\
MultiSentiNet-M\cite{Xu2017mvsa} & 68.2 & {\scriptsize (+8.4)} & 67.7 & {\scriptsize (+8.2)} & 64.2 & {\scriptsize (+14.2)} & 59.6 & {\scriptsize (+18.7)} \\
Co-Memory-M\cite{Xu2018Co-MemoryNetwork} & 71.1 & {\scriptsize (+5.5)} & 70.1 & {\scriptsize (+5.8)} & 64.3 & {\scriptsize (+14.1)} & 59.1 & {\scriptsize (+19.2)} \\
Se-MLNN\cite{Che2021Se-MLNN} & 73.1 & {\scriptsize (+3.5)} & 71.3 & {\scriptsize (+4.6)} & 58.2 & {\scriptsize (+20.2)} & 58.0 & {\scriptsize (+20.3)} \\
MGNNS\cite{yang-etal-2021-multimodal} & 71.0 & {\scriptsize (+5.6)} & 69.7 & {\scriptsize (+6.2)} & 65.3 & {\scriptsize (+13.1)} & 65.0 & {\scriptsize (+13.3)} \\
\hdashline
MVAN-M \cite{36_Tameemi2023mvan} & 73.0 & {\scriptsize (+3.6)} & 73.0 & {\scriptsize (+2.9)} & 66.5 & {\scriptsize (+11.9)} & 63.4 & {\scriptsize (+14.9)} \\
CLMLF \cite{06_li-etal-2022-clmlf} & 75.3 & {\scriptsize (+1.3)} & 73.5 & {\scriptsize (+2.4)} & 71.1 & {\scriptsize (+7.3)} & 71.0 & {\scriptsize (+7.3)} \\
CTMWA \cite{21_Zhang2024CTMWA}  & 75.9 & {\scriptsize (+0.7)} & 75.7 & {\scriptsize (+0.2)} & 68.6 & {\scriptsize (+9.8)} & 68.6 & {\scriptsize (+9.7)} \\
VSA-PF \cite{23_Chen2024Holistic} & 75.6 & {\scriptsize (+1.0)} & 74.5 & {\scriptsize (+1.4)} & 76.6 & {\scriptsize (+1.8)} & 76.6 & {\scriptsize (+1.7)} \\
\hdashline
BERT-ViT-EF (Ours) & 74.3 & {\scriptsize (+2.3)} & 73.3 & {\scriptsize (+2.6)} & 77.4 & {\scriptsize (+1.0)} & 77.4 & {\scriptsize (+0.9)} \\
\hline
\textbf{DTCN} & \textbf{76.6} & & \textbf{75.9} & &\textbf{78.4} & & \textbf{78.3} &\\
\hline
\end{tabular}
\label{tab:results_baseline_transform}
\end{table}
\vspace{-3mm}

Second, the addition of a Transformer encoder layer after BERT improves the contextual richness of text features. This refinement is not implemented in BERT-ViT\_EF, which uses BERT outputs directly without further contextual enhancement. Our ablation study confirms that increasing the number of self-attention layers to 1 yields a gain in accuracy and F1-score on MVSA-Single, highlighting the importance of deeper semantic understanding in the textual domain. 

Finally, the incorporation of contrastive learning provides a strong regularization signal that aligns paired text and image embeddings in the latent space. By enforcing proximity between matched cross-modal pairs and separation between unmatched ones, the model learns modality-invariant representations that generalize better to downstream classification. Unlike BERT-ViT\_EF, which lacks contrastive alignment, DTCN benefits from the synergy between refined textual encoding and cross-modal contrastive alignment enhances both robustness and discriminative power of the joint sentiment representation.

\section{Conclusion and Future Work}
\label{sec:conclusion}

In this paper, we presented DTCN, an MSA model that integrates BERT and ViT through an early fusion strategy. Additionally, DTCN incorporates an extra Transformer encoder layer after BERT to refine the textual context before fusion and employs contrastive learning to enable effective cross-modal interaction within a streamlined architecture. Extensive experiments on two benchmarks—MVSA-Single (4,511 text-image pairs, 3 classes) and TumEmo (195,265 pairs, 7 classes)—show that DTCN achieves higher accuracy (78.4\%) and F1-score (78.3\%) on TumEmo, with competitive results on MVSA-Single (76.6\% accuracy, 75.9\% F1).

The DTCN model improves accuracy and F1-score by a minimum of 1\% on MVSA-Single and outperforms existing methods by approximate 2\% on TumEmo. The additional Transformer encoder layers applied to BERT enhance textual representations by capturing long-range dependencies and subtle sentiment nuances, while contrastive learning aligns image-text embeddings to improve robustness and generalization. These results underscore the advantages of early fusion for jointly learning emotional cues from both modalities and represent significant progress over previous Transformer-based models. Collectively, these architectural and training innovations establish DTCN as a powerful and extensible framework for MSA, setting a new benchmark for future research in the field.

In future work, we aim to extend the proposed model to incorporate audio and video modalities for multimodal emotion recognition. Additionally, we plan to investigate lightweight encoder architectures (e.g., DistilBERT, MobileViT) to enable real-time deployment in resource-constrained environments. Furthermore, we will explore the integration of ontologies \cite{cao2012semantic}, knowledge graphs \cite{ngol2021semantic},  and data warehouse frameworks \cite{ngo2019designing,ngo2020data} to enhance both classification performance and computational efficiency.



\vspace{3mm}\noindent
\textbf{Funding:} This publication has emanated from research conducted with the financial support of Taighde Éireann – Research Ireland under Grant number 12/RC/2289\_P2. \vspace{8mm}

%

\begingroup  
\let\clearpage\relax  

\bibliographystyle{splncs04} 
\bibliography{ref}    
\endgroup
\end{document}